\documentclass[letterpaper, 10 pt, conference]{ieeeconf}
\IEEEoverridecommandlockouts                              
\usepackage{cite}
\usepackage{amsmath,amssymb,amsfonts}
\usepackage{algorithmic}
\usepackage{textcomp}
\usepackage{xcolor}
\usepackage[pdftex]{graphicx}
\graphicspath{{images/}}
\DeclareGraphicsExtensions{.pdf,.jpeg,.jpg,.png}
\usepackage{mathptmx} 
\usepackage{times} 
\usepackage{amsmath} 
\usepackage{amssymb}  
\usepackage{hyperref}
\usepackage{gensymb}
\usepackage{todonotes}
\usepackage[product-units=single,per-mode=symbol,range-units=single,range-phrase=\,--\,,detect-all]{siunitx}
\usepackage{balance}

\def\BibTeX{{\rm B\kern-.05em{\sc i\kern-.025em b}\kern-.08em
    T\kern-.1667em\lower.7ex\hbox{E}\kern-.125emX}}

    \title{\LARGE \bf Lessons from Robot-Assisted Disaster Response Deployments \\ by the German Rescue Robotics Center Task Force} 

\author{Hartmut Surmann$^{1}$, Ivana Kruijff-Korbayov\'a$^{2}$, Kevin Daun$^{3}$, Marius Schnaubelt$^{3}$, Oskar von Stryk$^{3}$,\\ Manuel Patchou$^{4}$, Stefan Böcker$^{4}$, Christian Wietfeld$^{4}$,  Jan Quenzel$^{5}$, Daniel Schleich$^{5}$, Sven Behnke$^{5}$, \\ Robert Grafe$^{6}$, Nils Heidemann$^{6}$, Dominik Slomma$^{6}$
\thanks{$^{1}$University of Applied Science, Gelsenkirchen (WHS), Germany;
$^{2}$German Research Center for Artificial Intelligence (DFKI), Saarbr\"ucken, Germany; 
$^{3}$Simulation, Systems Optimization and Robotics Group, TU Darmstadt (TUDA), Germany; 
$^{4}$Communications Network Institute, TU Dortmund (TUDO), Germany; 
$^{5}$Autonomous Intelligent Systems, University of Bonn (UBO), Germany;
$^{6}$German Rescue Robotics Center (DRZ)
}
\thanks{Corresp. author: {\tt\small hartmut.surmann@w-hs.de}}
}

\begin{document}

\maketitle


\begin{abstract}
Earthquakes, fire, and floods often cause structural collapses of buildings. The inspection of damaged buildings poses a high risk for emergency forces or is even impossible, though.
We present three recent selected missions of the Robotics Task Force of the German Rescue Robotics Center, where both ground and aerial robots were used to explore destroyed buildings.
We describe and reflect the missions as well as the lessons learned that have resulted from them.
In order to make robots from research laboratories fit for real operations, realistic test environments were set up for outdoor and indoor use and tested in regular exercises by researchers and emergency forces. Based on this experience, the robots and their control software were significantly improved. Furthermore, top teams of researchers and first responders were formed, each with realistic assessments of the operational and practical suitability of robotic systems.
\end{abstract}

{\bf Keywords}: Rescue Robotics, Ground and Aerial robots, Public Safety, Search and Rescue, Human-Robot Interaction

\section{Introduction}
\label{sec:introduction}



The use of robots, specifically unmanned ground and aerial vehicles (UGVs and UAVs), in situations involving structurally compromised buildings has obvious potential for increasing operational capability while maintaining personal safety of first responders. Structural collapse is characterized by the need for collecting data in places inaccessible using standard equipment and/or in environments risky to enter for humans. For example, in a pioneering deployment of robots in the earthquake-struck Emilia Romagna, Italy, 2012, a UAV provided exterior visual information of a church tower for structural damage assessment, and a UGV was used to explore the interior of a dome to assess the state of important religious artefacts~\cite{6523866}. In Amatrice, Italy, 2016, robots were used to provide detailed exterior and interior 3D models for the planning of shoring operations of two churches severely damaged by an earthquake~\cite{Kruijff-amatrice}. In both cases, robotic research demonstrators were used by academic research teams embedded within onsite fire brigade forces---enabled by existing collaborations between academics and first responders in research projects. 


This successful long-term collaboration model was incorporated into the setup of the German Rescue Robotics Center (DRZ)~\cite{9597869}, a non-profit association connecting academia, end users, and industry 
to facilitate the advancement and transfer of robotic technologies for first responders. DRZ conducts extensive field tests and has established a Robotics Task Force (RTF), which operates together with the Dortmund fire brigade (FwDo) and consists of professional fire fighters and researchers. 
In DRZ RTF deployments, robots are operated by researchers so far. This enables the use of cutting-edge robotic technology, including research demonstrators, for which the fire brigade does not yet have training. The RTF can, thus, evaluate the benefits and shortcomings of novel technology and increase awareness of its potential with first responders. We believe that such collaboration is crucial for advancement, both for determining a relevant research agenda and for driving innovation from the end-user perspective.


\section{Related Work}





\begin{figure}[!t]
\centering
\includegraphics[width=0.4\textwidth]{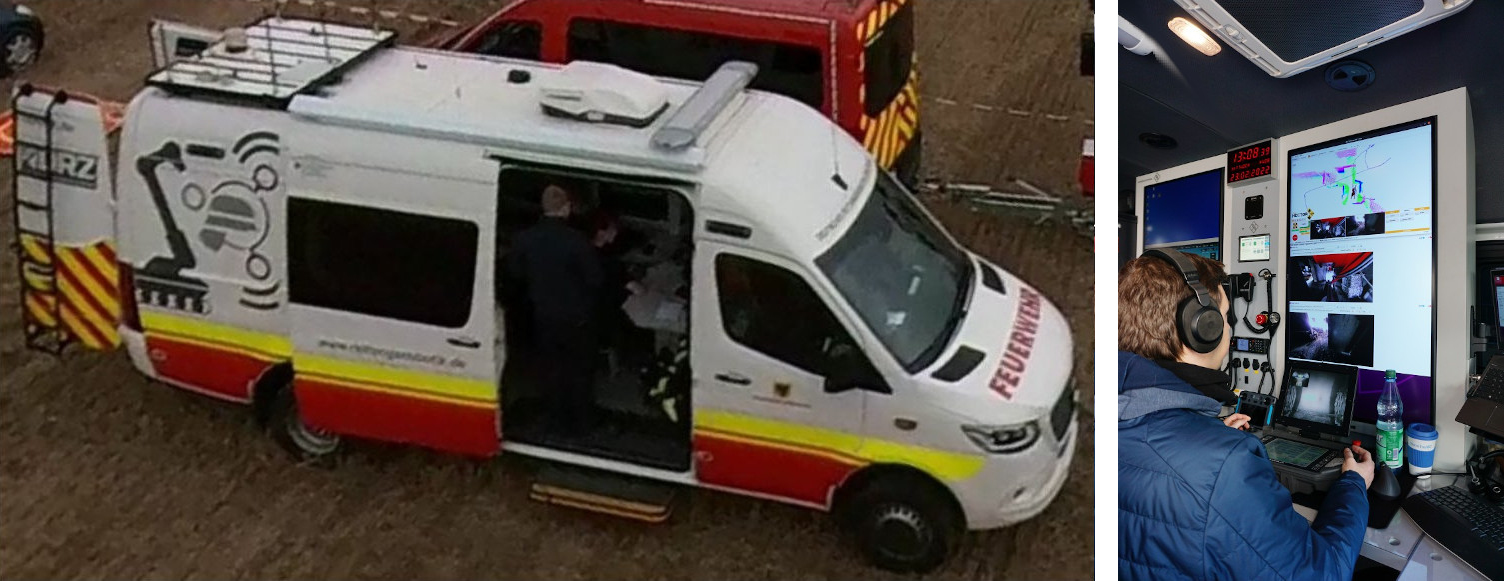}\vspace*{-1ex}
\caption{Overview of the RobLW. Left: The RobLW. Right: A robot operator in the command center with two monitor workstations.}
\label{fig:rlw-overview}
\end{figure}

Urban search and rescue and/or disaster response has high potential for the use of robots.  
An analysis of 114 calls completed by the Boulder Emergency Squad from 04/2016 to 12/2021 using unmanned aerial systems reveals the divergence in the assumptions made in research and how state-of-the-art technologies may realistically transition into operational capacity~\cite{OperationalUseofUAS}. Therefore, we focus on systems that are tested with end users and actually used in real deployments. 
CRASAR\footnote{\url{https://www.crasar.org}} was the first research team who started carrying out real deployments in 2002.
Murphy 
comprehensively summarized the early experiences~\cite{murphy2014disaster}.
Tadokoro et al. also have a variety of real world missions, especially in Fukushima~\cite{tadokoro2019disaster}. 
Some research projects with end users made it to the real world~\cite{6523866,Kruijff-amatrice}, 
while others remained singularities~\cite{DeCubber/etal:2013, Marconi/etal:2013}. 
Competitions such as Eurathlon, RoboCup or DARPA challenges focus on testing robotic capabilities, 
and not on embedding the robotic teams within realistic deployment conditions and  command structures.  
Fire fighters are increasingly employing UAVs \& UGVs, e.g, the mission at Notre Dame, 2019~\cite{nardi_2019_notredame}. 
They strive to build  their own expertise, often in specialized units.

\section{Deployment Setup of the DRZ RTF}



\subsection{Robotic Command Vehicle (RobLW)}


For in-field testing and even more during real missions, it is crucial to provide basic logistics, communication, and support for the teams and their robots, to not rely on already heavily occupied civil infrastructure.
For these purposes, the DRZ developed the Robotics Command Vehicle (RobLW), a fully equipped emergency vehicle with radio communication, a signaling system, and basic emergency equipment (Fig.~\ref{fig:rlw-overview}). RobLW can carry multiple UAVs and one mid-sized UGV, is able to set up network communication infrastructure for the RTF, and has a  map-based situation awareness system for mission control.

In its center, two fully equipped workplaces for a team commander and/or robot operator(s) for steering the robots and data management have been set up. They are embedded in an own network and server/ client infrastructure of the van, enabling communication (WiFi, Internet, dedicated robot communication) and data processing (e.g. calculation of 3D maps).
The rear area of the van offers storage capacity for various peripheral equipment and components. Furthermore, the compartment provides a transport storage area for robots. On the roof, a flexible antenna array is set up.  


One example for data processing is the usage of a Web\-ODM Server\footnote{Drone Mapping Software: \href{www.opendronemap.org/webodm/}{webodm}}, a web-based tool that uses camera images provided by UAVs to create (offline) a three-dimensional representation of the environment. Web\-ODM provides a scaled point cloud that can be used to measure the area of operation or holes in a building at risk of collapse. This facilitates situation assessment and approach planning~\cite{DBLP:journals/corr/abs-1709-00587}.

\subsection{Robotic Systems}


\paragraph{Unmanned aerial vehicles (UAVs)}
\label{sec:D1}

The DRZ RTF uses various drones of two distinct classes. 

The first class contains off-the-shelf commercial drones with proprietary remote controller, software, and radio communication provided by the respective manufacturer, e.g. DJI. 

\begin{figure}
\centering \footnotesize
a)~\includegraphics[width=0.4\textwidth]{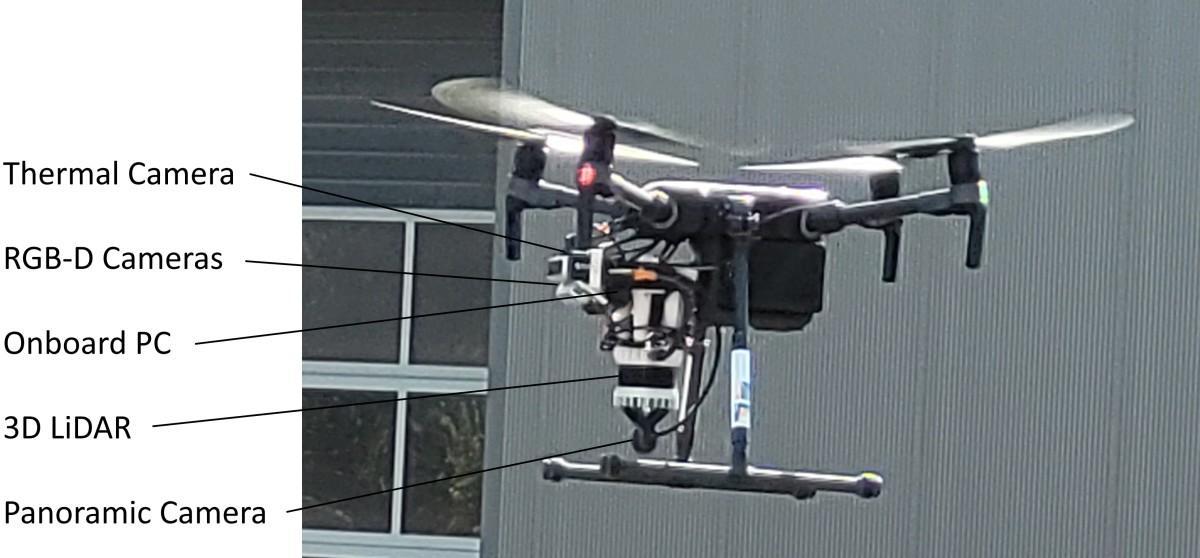}\vspace*{1ex}\\

b)~\includegraphics[width=0.4\textwidth]{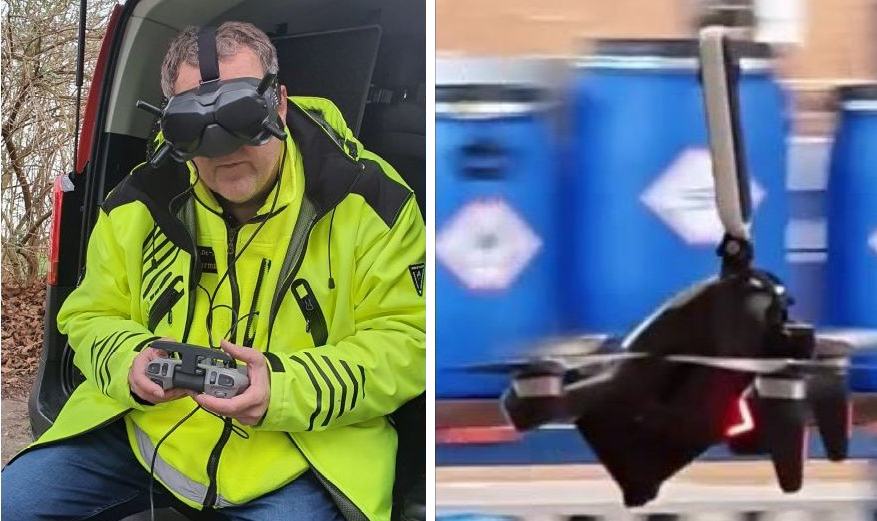}\vspace*{-1ex}
\caption{Eamples of UAVs used in DRZ. a) DRZ D1 Copter equipped with rich sensors and PC~\cite{Bultmann:ECMR2021}; b) DJI-FPV extended with a 360° camera.}
\label{fig:DRZ-Copter}
\end{figure}

The second class contains DIY and modified drones. These can carry different loads and are customized for specific uses. An example is the D1 Copter shown in Fig.~\ref{fig:DRZ-Copter}a. It is based on the DJI Matrice 210 v2 platform and is equipped for onboard environment perception and navigation planning~\cite{Schleich:ICRA2021} with an Intel NUC8i7BEH computer, allowing continued operation even during short communication outages.
An Ouster OS0-128 LiDAR enables 3D-SLAM and all-around obstacle avoidance~\cite{Schleich:IROS2022} as well as waypoint navigation and exploration with minimal effort from the operator using a gamepad controller. The NUC's iGPU runs CNN inference for semantic segmentation and person detection from point cloud as well as color and thermal imagery~\cite{Bultmann:ECMR2021,Bultmann:RAS2022}.
For mission control and operator supervision, relevant status information and preprocessed measurements are transmitted over WiFi to a ground station, where the live reconstructed 3D color map is visualized along with images from an Insta360 Air panoramic camera, a FLIR ADK thermal camera, and semantic information.

\paragraph{UGV with manipulation ability} 
The RTF uses a UGV with grasping capability (see Fig.~\ref{fig:Essen}) as exploration and manipulation platform.
A Telerob Telemax Hybrid is equipped with multiple modules to enable operator support with assistance functions, such as 3D-SLAM\cite{daun2021}, obstacle avoidance, waypoint navigation as well as autonomous exploration.
The splash-proof navigation module mounted on the back of the UGV provides perception by combining a continuously rotating LiDAR, an omnidirectional camera, and multiple RGB-D cameras. The sensor module at the gripper consists of a thermal camera, a RGB-D camera, a HDR wide-angle camera, and a zoom camera. The sensor data is processed on the robot using onboard-computing and transmitted via WiFi connection to the operator, who can control and supervise the assistance functions.

\begin{figure}
\centering \footnotesize
a)~\includegraphics[width=0.4\textwidth]{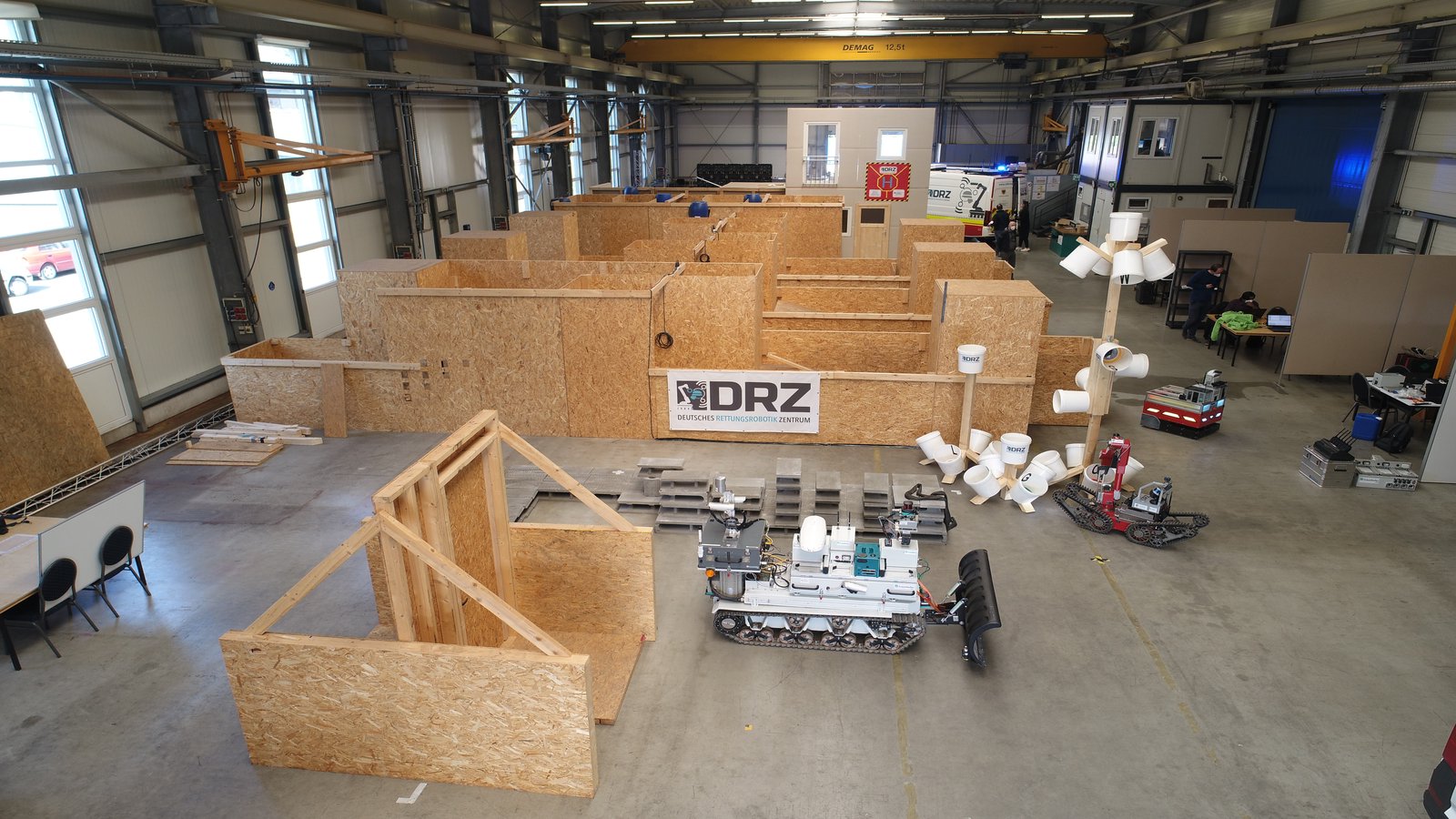}\vspace{1ex}\\
b)~\includegraphics[width=0.4\textwidth]{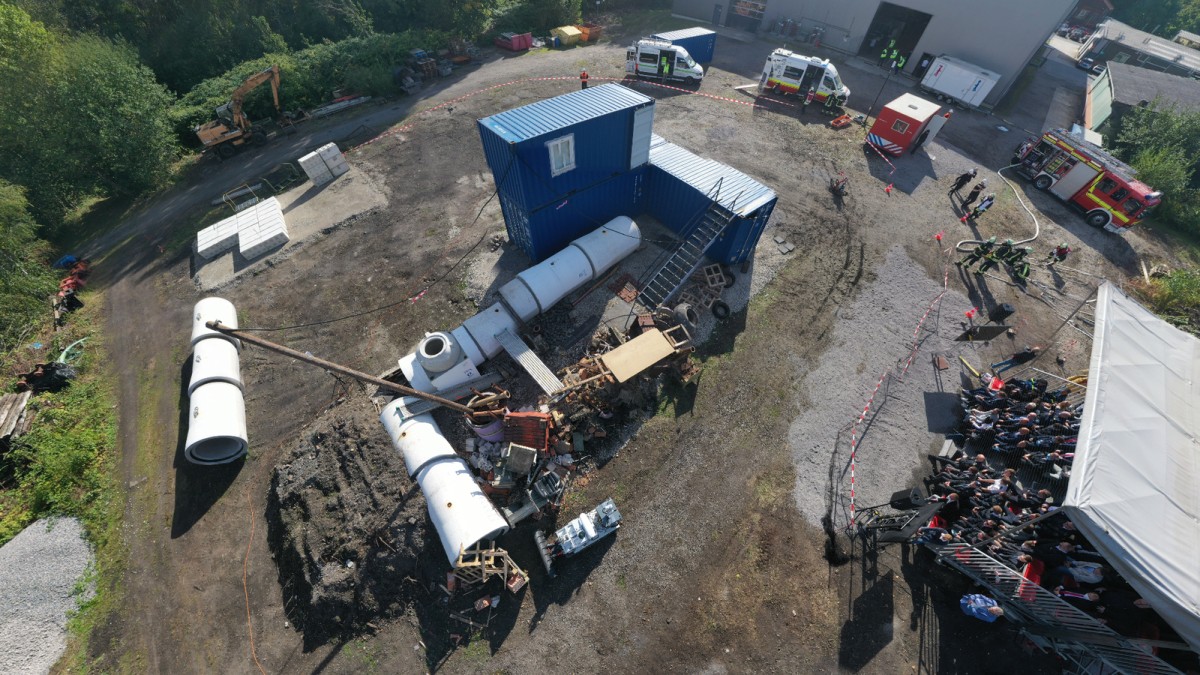}\vspace*{-1ex}
\caption{DRZ training facilities. a) Training's hall with NIST UAV and UGV training ground; b) outside training ground with container and ruble.}
\label{fig:drz-environment}
\end{figure}

\subsection{Network Communication Platform}

Despite state-of-the-art robotic systems already being able to perform with partial or full autonomy, reliable communications remain indispensable for mission configuration and monitoring and for emergency-related real-time teleoperation. Since rescue missions usually exhibit challenging network conditions, such as increased signal attenuation through multiple layers of collapsed walls and electromagnetic interference from malfunctioning devices or damaged power lines, resilient networking solutions are required.

\begin{figure}
\centering
\includegraphics[width=0.4\textwidth]{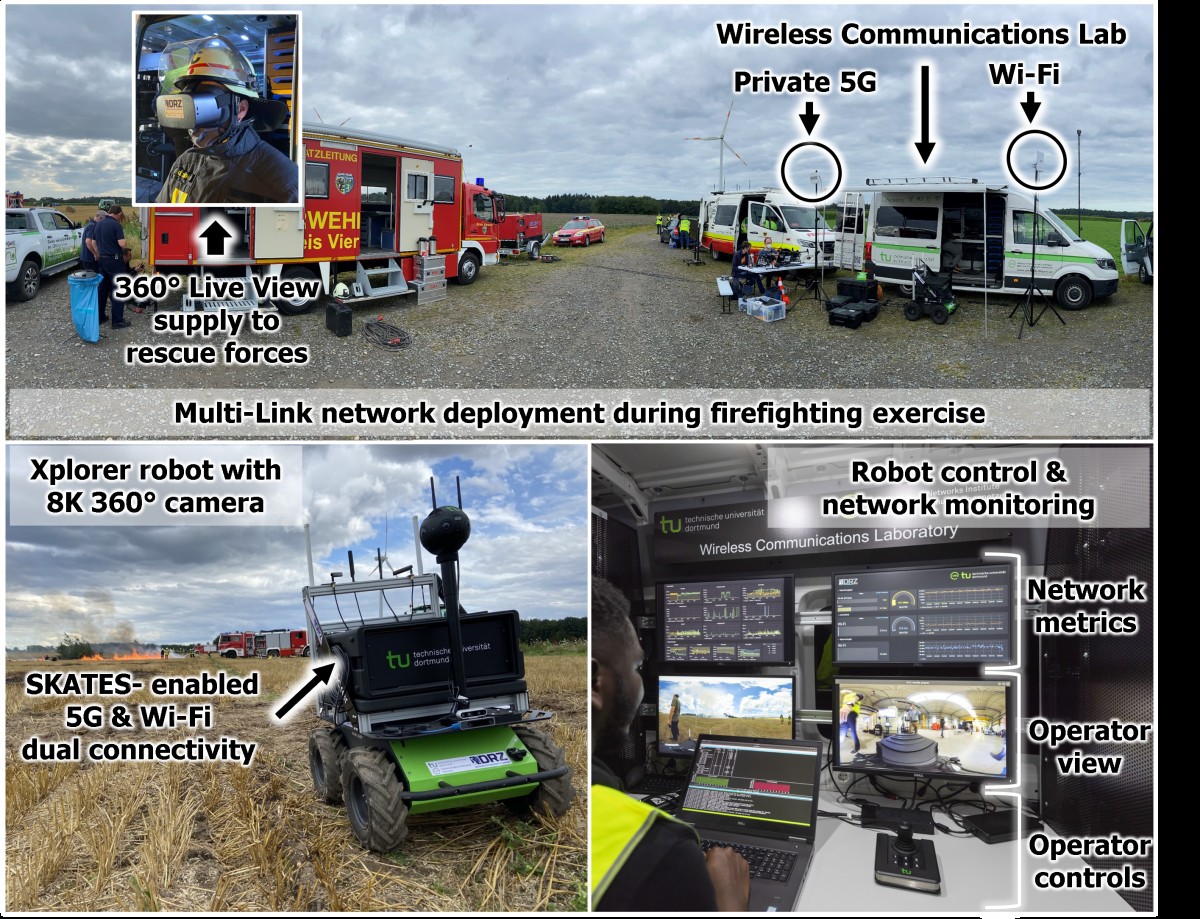}\vspace*{-1ex}
\caption{Deployment of interoperable Multi-Link communication platform designed for DRZ missions, provisioned with 5G and Wi-Fi during a firefighting exercise to supply immersive situational awareness to the rescue forces.}
\label{fig:netcomms-platform}
\end{figure}

The SKATES\cite{Gueldenring2020} communication platform was developed and integrated in the DRZ networking approach to provide a robust and interoperable means of acquiring sensor data, transmitting steering commands, and enabling real-time multimedia supported mission orchestration. Through the interoperable nature of the SKATES module's connectivity, a flexible blend of Radio Access Technologies (e.g Wi-Fi, 4G, 5G, or IP-Mesh networks) is enabled through a multi-connectivity approach to improve the overall robustness of the communication link. SKATES was tested on various occasions, such as the firefighting exercise organized in Viersen, Germany, 2021 (Fig.~\ref{fig:netcomms-platform}), where it was provisioned with Wi-Fi and 5G connectivity, thereby extending the robot's range to reach deeper parts of the field. More details to the SKATES platform are provided in \cite{9597869} and \cite {Gueldenring2020}.

\subsection{Joint Exercises in Living Lab and Realistic Scenarios}  

Deploying research demonstrators in actual disasters requires preparation and training.
Within the A-DRZ project, we performed complex, close to realistic scenario tests (Fig. \ref{fig:drz-environment}) together with professional first responders every 6 months, to train, refine requirements, evaluate current solutions and build up team experiences.

RobLW is shared by the DRZ RTF---for research---and 
FwDO---to test it in real deployments. The DRZ RTF setup is regularly tested in joint exercises in the DRZ living lab~\cite{9597869} and ready to deploy 24/7. 


\section{Deployment Reports}\label{sec:deployment-use-cases}

\subsection{Industrial Hall Fire Berlin}
\label{sec:berlin}
\begin{figure}
\centering
\includegraphics[width=0.4\textwidth]{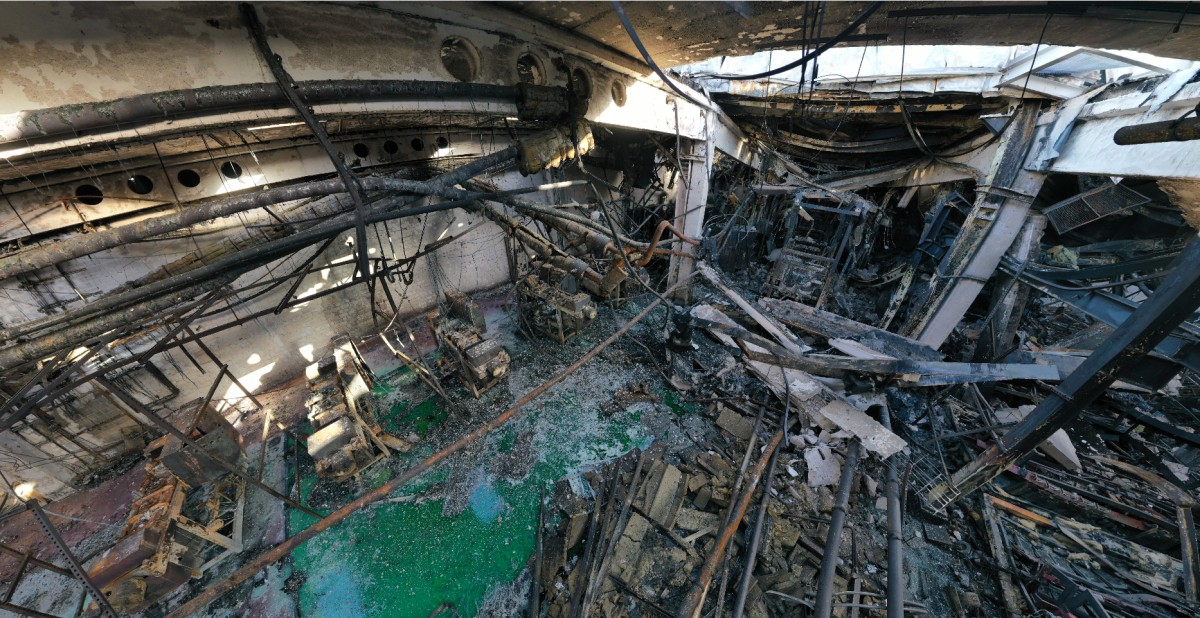}\vspace*{-1ex}
\caption{Snapshot out of an UAV made panorama after the fire. The green substance is highly toxic cyanide. The metal parts and cables hanging around make flying extremely difficult with a correspondingly high risk of losing the drone \cite{9597677}. }
\label{fig:Berlin}
\end{figure}

\noindent$\circ$~\textit{Situation:} 
On Feb.\,11\textsuperscript{th}, 2021 a fire broke out in a metalworking factory in Berlin, Germany, and could only be extinguished after more than 12\,h of work.
Hazardous substances were released during the fire 
(Fig.~\ref{fig:Berlin})\footnote{\href{https://www.berliner-feuerwehr.de/aktuelles/einsaetze/grossbrand-in-stoerfallbetrieb-in-berlin-marienfelde-3719}{Grossbrand Berlin}}.
Due to the high level of damage, an entry ban was issued. 
On Feb.\,22\textsuperscript{nd}, the Berlin police submitted an administrative assistance request to the Dortmund fire department for support with special UAV technology as part of fire investigation. 

\noindent$\circ$~\textit{Team composition:} A team of emergency responders from FwDO, DRZ staff members, and WHS researchers was put together and set off to Berlin for a three-day mission with the RobLW. The roles needed in the team were:
team leader, UAV pilot(s), and IT expert(s).

\noindent$\circ$~\textit{The task} for the team was to build a digital representation of the outside and inside area of the industrial hall.
The RobLW was used on site to process the UAV images to create 3D views while displaying live-streamed footage.
Due to the remaining 10-30\,cm high and potentially contaminated extinguishing water on the ground, it was clear from the outset that ground robots could not be used.

\noindent$\circ$~\textit{Mission execution:}  
Since the hall was unknown terrain, first an overview was needed. For this purpose, a meander flight with a commercial DJI Mavic 2 was made 45\,m above the hall. The downward-facing UAV images were processed after landing to an orthophoto and a georeferenced 3D point cloud model of the hall. The model was then used to measure and survey windows and openings for possible entry.

From the openings measured above, it was obvious that flying through the ceiling opening into the hall was possible with a Phantom 4 and a Mavic 2, but two issues needed to be addressed. The small aperture angle of the UAV camera (FoV $\approx$ 80°) did not allow seeing the boundaries of the opening while flying through it. The second issue was to fly the UAV back out of the hall, because the UAV cameras could be pointed forward and downward but not upward. To ensure visibility, we took the \textit{two pilots approach}. When the first UAV enters the hall, the second UAV is positioned exactly above the entry point and assists in flying in. While the first UAV autonomously creates the panoramas in the hall, the second UAV waits above the opening, and after completing the shots, the second pilot navigates the first UAV back out through the opening. To fly in, a Mavic 2 was chosen due to the slightly better camera resolution and was equipped with propeller guards. Upon approaching the openings, it was noticeable that the metal of the roof hatches had been severely bent by the fire. This significantly reduced the width of the holes and made it very difficult to fly into the hall. Some openings could not be flown at all. 

Another trick was used to capture data from these remaining obstructed positions. 
A panoramic camera, Insta360 One X (110\,g, 15.8\,MP), was attached to the Phantom 4 with a 1.5\,m long thin rope and inserted into the respective opening from above. With this, it was possible to record 20\,min of video to view all parts of the site.

\noindent$\circ$~\textit{Lessons learned:} Flying in a hall heavily destroyed by fire is extremely risky and difficult, but the recordings of the high quality images (5.7k)---especially the panorama images (16k)---as well as video material\footnote{UAVs Berlin Video: \href{https://youtu.be/mR05-akD4BE}{ https://youtu.be/mR05-akD4BE}} (4k) was vary helpful for the situation awareness. The material was made available to the Berlin police for assessing the situation and for further investigations. Furthermore, setting up appropriate environments and flying in them must be trained over and over again and incorporated into the fire department's training plan. Moreover, easy-to-fly (autonomy, obstacle avoidance), much smaller drones ($<$ 30\,cm) with a 360° camera, and improved radio connectivity are helpful, especially in environments with much metallic shadowing. 

\subsection{Flooded Town Erftstadt}

\begin{figure}
\centering
\includegraphics[width=0.48\textwidth]{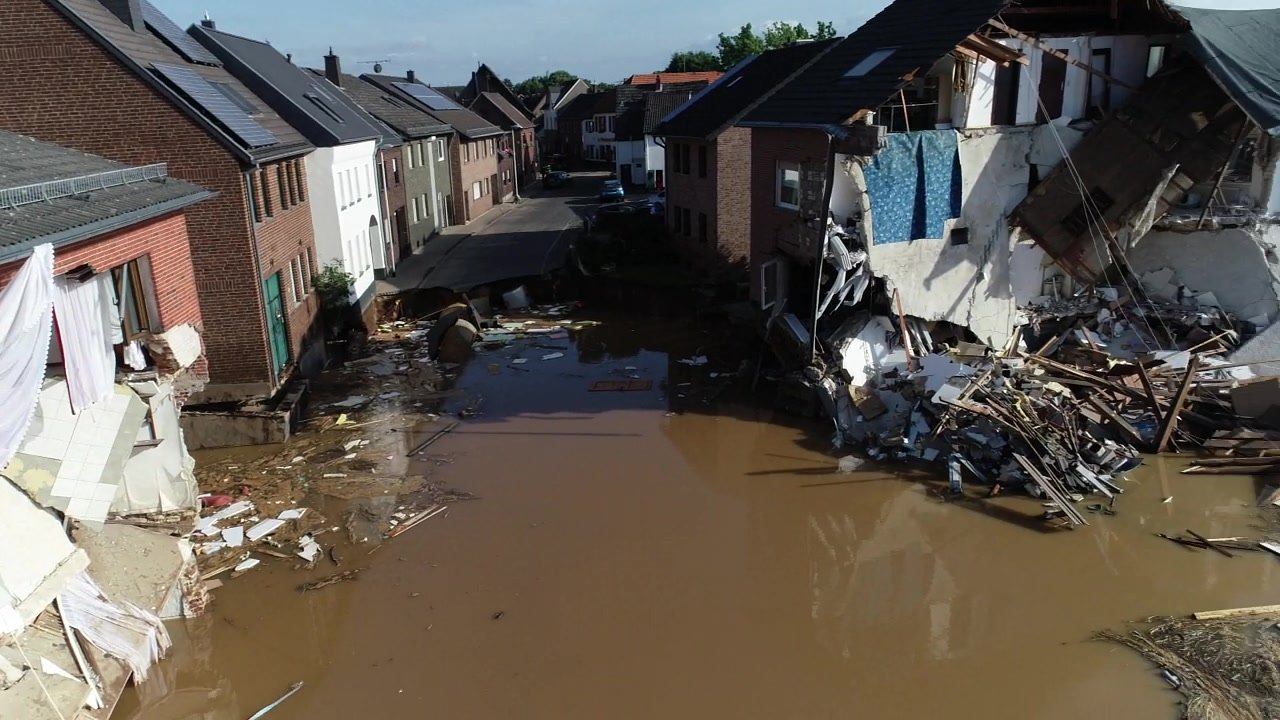}\vspace*{-1ex}
\caption{Urban flooding, overview, person search, Erftstadt / Blessem, Germany 2021~\cite{9738529}. }
\label{fig:Erftstadt}
\end{figure}

\noindent$\circ$~\textit{Situation:} The Flood in Western and Central Europe in July 2021 was a natural disaster with severe flash floods in several river basins (Fig.~\ref{fig:Erftstadt}). Parts of Belgium, the Netherlands, Austria, Switzerland, Germany, and other neighboring countries were particularly affected. The most severe floods were caused by thunderstorm Bernd\footnote{2021 European floods: \href{https://en.wikipedia.org/wiki/2021_European_floods}{https://en.wikipedia.org/wiki/2021\_European\_floods}}.
The drastic consequences of the storm disaster in Western Germany also made themselves strongly felt in Erftstadt / Blessem. Due to the flooding and a potential collapse of the Steinbach dam, thousand residents in several localities had to be evacuated from their homes. The Erft and Swist rivers had burst their banks and flooded large parts of the Erftstadt urban area. Long-distance roads such as the federal highways No.~1 (Eifelautobahn) and No.~61 as well as the federal highway 265 were closed as a result of the flooding and road damage. In the Erftstadt district of Blessem, the waters of the Erft flowed through a residential and commercial area and made a new path into the pit of the Blessem gravel plant; several houses were washed out, several others damaged near Blessem Castle.
An extensive emergency response mission commenced, including a rescue robotics team of the DRZ.

\noindent$\circ$~\textit{Team composition:} The team consisted of personnel from FwDO, DRZ, UBO, UzL, TUDA, and WHS. It set off to Erftstadt/Blessem for a two-day mission with the RobLW directly in this very dangerous area at Blessem Castle. The roles needed in the team were: team leader, UAV pilot(s), and IT expert(s).

\noindent$\circ$~\textit{The tasks:}
\begin{enumerate}
    \item Live air observation of the demolition edge and alerting in case of further demolition or changes, especially to secure the emergency forces in search for missing persons in buildings behind the demolition edge,
    \item Generating high-resolution 3D models for the purpose of further mission planning directly on site even without power, internet, or mobile phone connection, 
    \item Detailed inspection of all buildings / structures that could not be accessed or reached by responders because many people were missing, and
    \item Creating clear and easily accessible documentation for the emergency services.
\end{enumerate}

\begin{figure}
\centering
\includegraphics[width=0.24\textwidth]{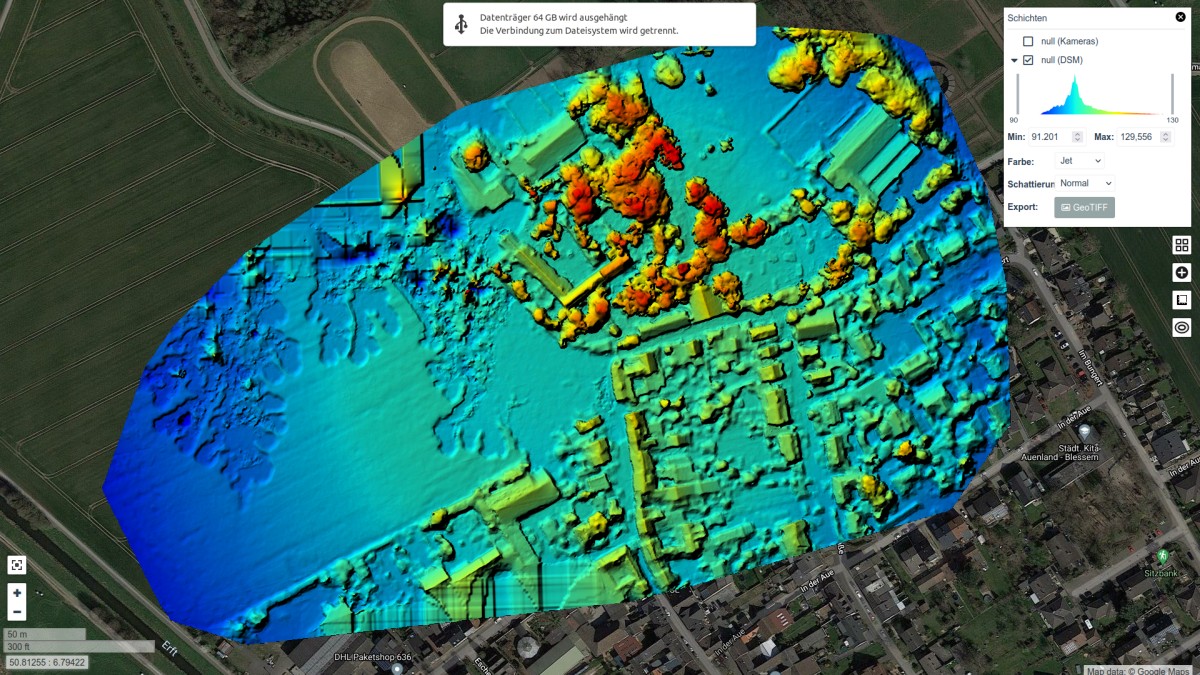}~\includegraphics[width=0.24\textwidth]{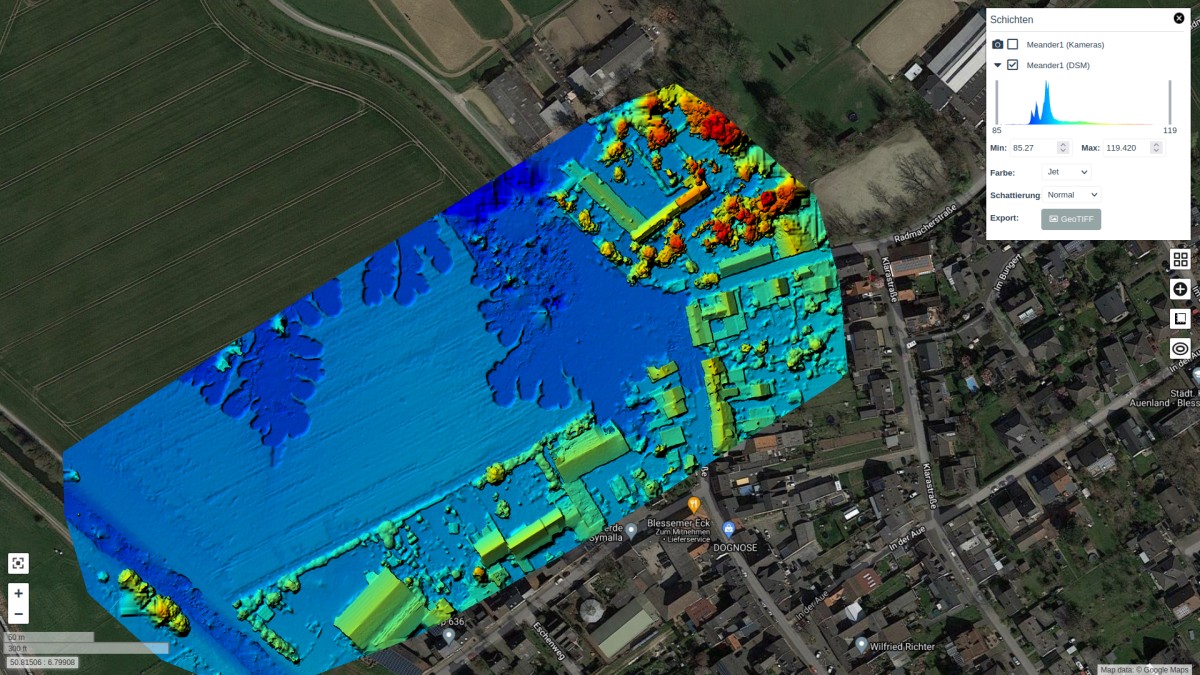}\vspace*{-1ex}
\caption{Elevation profiles of consecutive days. The comparison of the two elevation profiles shows that the water has sunk by about 40\,cm within one day, threatening further demolitions due to the lack of water backpressure. Based on the information, the emergency forces were pulled back by 100\,m from the demolition site.}
\label{fig:tiefenprofile}
\end{figure}

\noindent$\circ$~\textit{Mission execution:} Since the environment was destroyed over a large area, planning with existing information was not possible. So, first a drone (Yuneec Typhoon) was used to get live and overview images of the area and to record the extent of flooding and the direction the water flow. Second, systematic flights, i.e. meander flights, were planned and executed with commercial UAVs (Mavic 2, Phantom 4). From these images, a georeferenced orthophoto and 3D point cloud model was created. In addition to the meander flight, a 360° panorama was created by taking individual photographs in order to obtain an overview as quickly as possible.
Based on the orthophoto and the 3D point cloud model, the detailed investigation was planed and executed with a DJI FPV drone.
Especially the inaccessible and partially destroyed buildings and vehicles were inspected (Fig.~\ref{fig:Erftstadt}).
All information aggregated during the flights were compiled into a presentation and presented to the command staff later that night.
Due to the destroyed infrastructure (no electricity, no Internet), only the RobLW could be used for data processing on site, which was accordingly heavily utilized.
On the following day, further meander flights were performed, comparing the resulting 3D model and elevation profiles with those of the previous day (Fig~\ref{fig:tiefenprofile}). The resulting height difference of about 40\,cm shows a significant water runoff, which is why the protection zone along the edge of the break-off had to be enlarged again.

\noindent$\circ$~\textit{Lessons learned:} i) Large-scale emergencies like the flooding require the deployment and coordination of many UAV teams. 
ii) The failure of the infrastructure (power, Internet) necessarily requires the operation of power and Internet independent vehicles such as the RobLW. 
iii) Especially for large-scale damage events, the systematic aggregation of individual images into large high-resolution maps which can be overlaid with Google maps is particularly helpful to assess the current extent of destruction.
iv) The calculated georeferenced 3D points clouds and orthophotos are very suitable for further mission planning, especially the detection of critical locations (e.g. the break-off edge, partially destroyed houses) and for documenting subsequent detailed inspections with small FPV drones.
v) Small FPV drones with a large camera aperture angle are great for the detailed inspection of the collapsed houses.
vi) Due to the lack of experience, and thus the ability to use robots, on the part of the new first responders, the UGV and the larger UAV were not used. Deployments of the other UAVs were significantly injected by the scientists involved.


\subsection{Residential Complex Fire Essen}


\noindent$\circ$~\textit{Situation:} In the night to Feb.\,22\textsuperscript{nd}, 2022 a large residential apartment building in Essen, Germany caught fire.
Fanned by a storm, the fire spread quickly so that the entire southwestern facade was in flames. 
After the extinguishing work, 39 apartments on four floors were completely burned out.
Others were destroyed by smoke or extinguishing water. Due to the partly massive destruction, an entry ban was imposed. Although no one was missing after evaluation of the occupant numbers, the actual situation remained unclear due to the entry ban, which is why air and ground robots of the DRZ were requested.


\noindent$\circ$~\textit{Team composition:} The roles needed in the team were: team leader, UAV pilot(s), camera copilot(s), UGV pilot(s), safety officer(s), and IT expert(s).

\noindent$\circ$~\textit{The task} consisted of reconnaissance, clarification of the cause of the fire, and documentation of the scene. Small FPV drones ($<$ 1kg) with a 360° camera were deployed for the first time worldwide directly on Feb.\,22\textsuperscript{nd} for reconnaissance in the particularly heavily destroyed central part of the building complex and on Feb.\,23\textsuperscript{rd} ground and aerial robots were deployed in the less destroyed outer areas.

\vspace*{1mm}\noindent$\square$~\textit{Day I: FPV + 360° Reconnaissance}\\ 
As mentioned above, the \textit{task} on the first day was the exploration of the particularly destroyed areas, especially in the upper floors, which were no longer accessible due to the destruction of the stairs. For this purpose, the RobLW was additionally equipped with eight different drones and brought to the site. At the beginning of the \textit{mission}, a georeferenced 3D model of the operational environment was created by means of a 10\,min meander flight and subsequent 15\,min model calculation. The model was used to plan two FPV flights with a 360° camera. For example, a flight of 4:30\,min allowed five apartments to be completely examined. Videos of the flights and the created maps are available online\footnote{YouTube videos: 
\href{https://www.youtube.com/watch?v=Pd2__gm0nUE}{Essen360°}, 
\href{https://www.youtube.com/watch?v=iFE1kWW_jM4}{PanoViewer},
\href{https://www.youtube.com/watch?v=joXGfIUy2mc}{DenseMapping}
}.

\noindent$\circ$~\textit{Lessons learned:} FPV flights with a 360° camera create an impressive immersion. The small and lightweight drones can safely inspect collapsed buildings faster than humans, especially in the upper floors and severely damaged areas with non-existent staircases.

\begin{figure}
\centering
\includegraphics[width=0.48\textwidth]{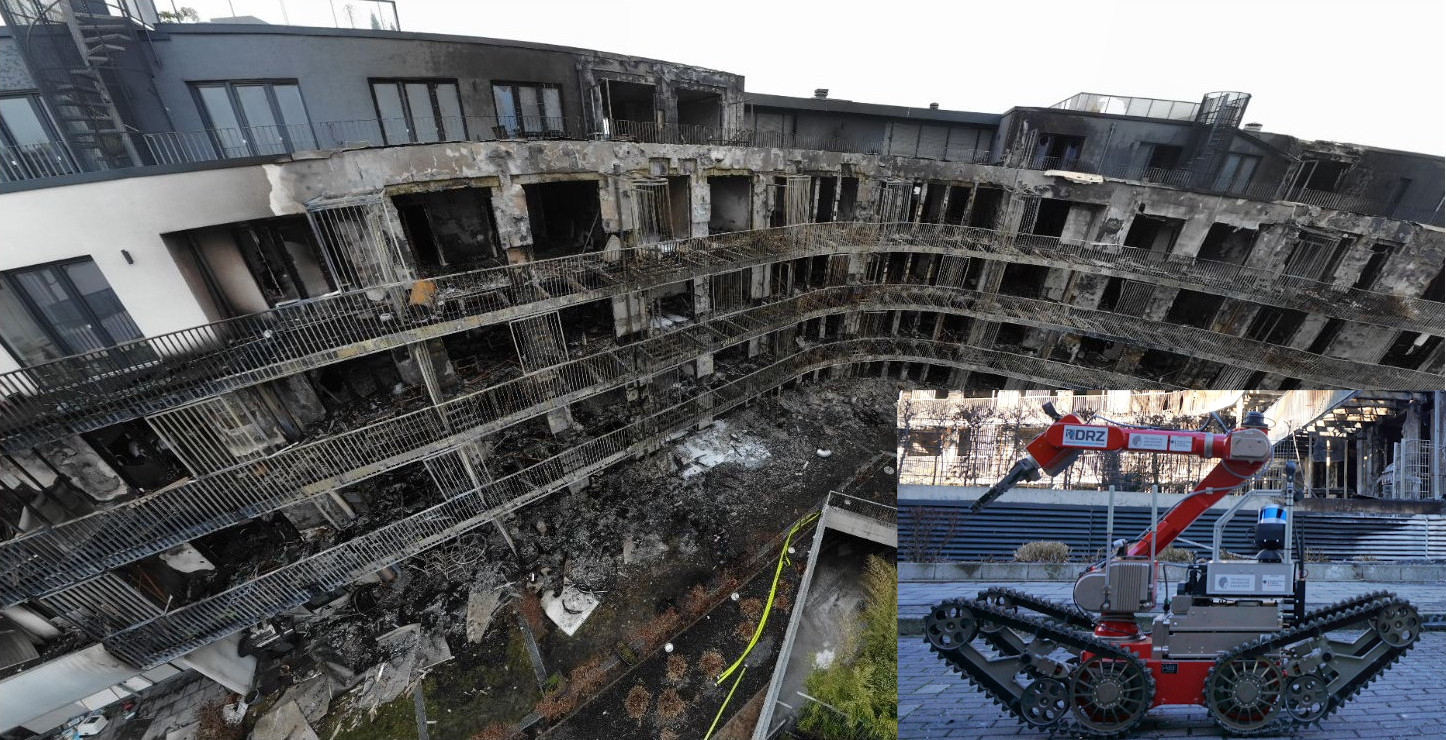}\vspace*{-1ex}
\caption{Residential fire, Essen, person search, cause of fire, Germany 2022. }
\label{fig:Essen}
\end{figure}

\vspace*{1mm}\noindent$\square$~\textit{Day II: Joint UGV/UAV deployment}\\ 
\noindent$\circ$~\textit{Task and setup:} The rapid spread of the fire was surprising for experts.
In order to avoid similar events at other buildings, the clarification of the cause of the fire was of high importance.
The Essen police submitted an administrative assistance request to the Dortmund fire department asking for site inspection support.
As a result, a team of emergency services from FwDO providing an UAV (DJI Matrice M300) and staff members from DRZ and TUDA with a tracked ground robot (see Figure \ref{fig:Essen}) were put together and set off to Essen for a one-day mission with the RobLW.
The \textit{task} was to create a digital 3D model and images of the inside of four flats within the vicinity of presumed origin of the fire. 

\noindent$\circ$~\textit{Mission execution:} 
After a meeting of the forces from the local police, FwDo, DRZ, and TUDA, a walk through the stable parts of the building was performed to assess the conditions.
The initial inspection helped to identify potential risks for the ground robot operation, such as narrow passages, large loose rubble, and wire meshes from burnt out couches.
Three inspections with the robotic systems were performed: One for each of the two flats in the first floor, which took roughly 45\,min each, and a third inspection of the two flats on the second floor, which took 75\,min.
To mitigate the risks of losing the robotic system during the exploration of the first floor, TUDA personnel had visual contact with robot from a safe distance for most parts of the mission and maintained radio contact with the robot operator.
Due to the concerns about the structural integrity, this was not feasible for the second floor.
Instead, a UAV was deployed to provide an outside camera perspective on the robot operation, which was shared with the UGV operator via a tablet showing the live video stream from the UAV.
After inspection, the lower part of the robot was covered with a hazardous dirt crust consisting of wet burnt ashes. Hence, the robot was decontaminated at the local fire department in Essen.\\ 
First 3D models of the explored environment could be computed live, during the mission.
To optimize the quality and provide tools to enable fire investigators to interact with the model, the data was processed again offline and submitted one week after the incident. 

\noindent$\circ$~\textit{Lessons learned:}\\
\noindent i) Robot mobility: The narrow shape of the ground robot allowed a deployment to most parts of the environment and to even pass through very narrow passages like a jammed door on the second floor.
The tracked drive worked well for traversing stairs, debris, and rubble. However, especially in curves caution was necessary as the tracks tended to dig into the loose ground.
The traversal of environment took rather long.

\noindent ii) Network connection: While driving inside the building with reinforced concrete walls and floors, the radio communication between the UGV and the robot operator was heavily dampened. This resulted in temporary WiFi connection loss with the robot, hindering the control of the assistance functions. However, the UGV is equipped with a second high-power proprietary radio connection which proved to be more reliable. 

\noindent iii) Assistance functions: Due to the challenging characteristics of the environment, robot operation was very challenging. The visualization of the registered point cloud~\cite{daun2021} with the 3D robot model in the user interface~\cite{fabian2021} strongly helped to navigate through narrow environments, as potential collisions with the surroundings could be precisely assessed. Furthermore, the rendering of virtual pinhole cameras~\cite{oehler2021} from the omnicamera helped to improve the situational awareness.
However, the availability of these assistance functions to the operator depended on the availability of the WiFi network, which was available inside the building for an estimated 70\,\% of the time with sufficient connection quality. In times without available WiFi connection and thus assistance functions, safely controlling the robot proved more challenging. 
Due to concerns about the performance of the control system in loose ground, no more complex assistance functions such as autonomous waypoint navigation were deployed.

\noindent iv) Multi-robot collaboration:
The deployment of an UAV to provide an outside perspective of the UGV for the exploration of the second floor was helpful to improve the situational awareness, especially in areas without available assistance functions due to poor WiFi connectivity. A drone equipped with a SKATES module acting as a movable relay may improve radio communication in future missions.

\noindent v) Robot robustness: Although developed as a research demonstrator, the developed splash water protection was necessary to perform the mission as extinguishing water was dropping from the ceiling and the system needed to be decontaminated after the mission.

\noindent vi) Team interaction: Mission control (professional first responders) and robot operation personnel (research staff) had joined exercises together in A-DRZ.
This helped to learn about communication, capabilities, and deployment procedures---enabling an efficient execution of the overall deployment. In our opinion, this is especially crucial for the participation of researchers without experience as first responders.

\noindent vii) Deployment preparation: As the deployed UGV is still under active development and a research demonstrator, extensive system checks at home and travelling to site on the day before deployment helped to mitigate risks. However, the trade-off between response delay and efficiency needs to be chosen for every mission.


\section{Discussion of Lessons Learned and Outlook}



Over the past 10 years, we have observed a shift in priorities regarding the usage of UAVs and UGVs for structural collapse inspection: at the beginning there was focus on livestream images (e.g., Emilia 2012~\cite{6523866}), then models were computed offline (e.g., Amatrice 2016~\cite{Kruijff-amatrice}), nowadays---thanks to technology advances---models can be computed in short time and used onsite for further mission planning, e.g., exterior models used for measurements (cf. Sec.~\ref{sec:berlin}). 
Regarding data products for post-mission analysis, such as detailed inspection of structural damage, there is a shift from 3D point cloud models to georeferenced, semantic 3D models and to localized high-resolution panorama pictures and further data processing with advanced photogrammetry and AI algorithms.
On the other hand, 3D point cloud models are very useful for teleoperation. 

UAVs are becoming smaller and are equipped with better cameras (IR, 360\degree, zoom) for delivering high-quality pictures ($>$\,4k). Flight modes are improving and provide more and more support for the pilot. This makes UAVs very suitable for reconnaissance, especially in structural collapse scenarios with tight, damaged conditions and rubble, which is difficult for UGV traversal. On the other hand, UGVs are generally better equipped for manipulation (e.g., opening doors, removing obstacles) and to carry varying payloads. Also, first responders note (p.c.) that air turbulence caused by UAV rotors poses a concern for spreading air pollution in some scenarios involving hazardous gasses. 

A typical mission process involves quick overview flights for initial planning and determination of ingress points, and multiple subsequent UAV and UGV sorties for detailed inspection and/or data collection.
This means that the remote presence and taskable agent roles defined by Murphy~\cite{murphy2014disaster} are often mixed, as data products are being used for problem solving on the spot, to determine what can be done during a mission at all, involving eye-inspection of both pictures and models. Consequently, there is a demand for quick creation of data products for mission planning. In the future, this needs to include map representations of trajectories and timeline calculations, integration of data from multiple sources, and dynamic projections for fast-changing situations. We plan such extensions of the Situation Awareness Interface described in \cite{9597869}. Further research on user interfaces is needed to appropriately support different tasks and roles in the team and communication between team members.  

Our experience regarding mission planning and task assignments is that the end users often do not know how to employ robots and what they can ask for in terms of available capabilities. The researchers involved in the DRZ RTF bridged this gap successfully by proposing possible actions and pro-actively contributing to the mission. Mission commanders who had previous exposure to UAVs/UGVs in action during joint exercises were more likely to request the RTF deployment.

We have repeatedly observed the need for multi-robot collaboration. On the one hand, 
we need to automate existing data capture and processing for larger areas with multiple small, affordable UAVs based on consumer platforms.
On the other hand,  UAV-UAV or UAV-UGV collaboration is needed to support teleoperation in situations where a primary UAV/UGV operator needs an external view (provided by a secondary UAV/UGV), e.g., for manipulation, entering and navigating in a (damaged) structure, traversing ruble, or passing through obstacles. The primary operator needs to be able to (verbally) coordinate with the secondary operator. He also needs a Camera Copilot to view the secondary video feed and provide additional guidance. Automation of the secondary operator's task is an interesting future research opportunity.   

Furthermore, more research is required to autonomously operate off-the-shelf UAVs not only in the vicinity of obstacles, but in confined spaces where loss is probable. Size and weight limitations mandate ground-based computation of live-streamed images and remitted control commands over wireless connections. The next important developments are needed in SW and AI, especially latency-aware perception and planning, fusion of local scene models, automatic assessment, better simulation (for planning and training), and handling changes in highly dynamic settings. Together, this will enable a small team of operators to configure and oversee a UAV swarm in real-time during continuous operation from a ground station.


The RTF team composition varied, depending on the task(s), whether both UAV(s) and a UGV were used, and the circumstances.  The following role distribution evolved with growing experience of the RTF: UAV/UGV pilot(s); camera copilot(s) in case of multi-robot collaboration; UGV safety officer for distant observation; IT expert(s) for data processing; and team leader, who is also the one to interface with the first responder corps. We did not use payload specialists~\cite{MurphyTadokoro2019}. During joint exercises in the DRZ living lab, we have also experimented with more complex teams using multiple UAVs and UGVs simultaneously~\cite{9597869}. In this case, we introduce a more hierarchical command structure, in which the UAVs are assigned directly to a mission commander, whereas UGVs form a group assigned to a group leader who reports to the mission commander. The integration of robotic (sub-)teams into the established first response command structures needs more work in the future. We need to specify ``blueprint" structures and roles for various types and sizes of missions, and include explicit planning of the command structure and role assignment in mission planning.

 

\begin{figure}
\centering
\includegraphics[width=0.48\textwidth]{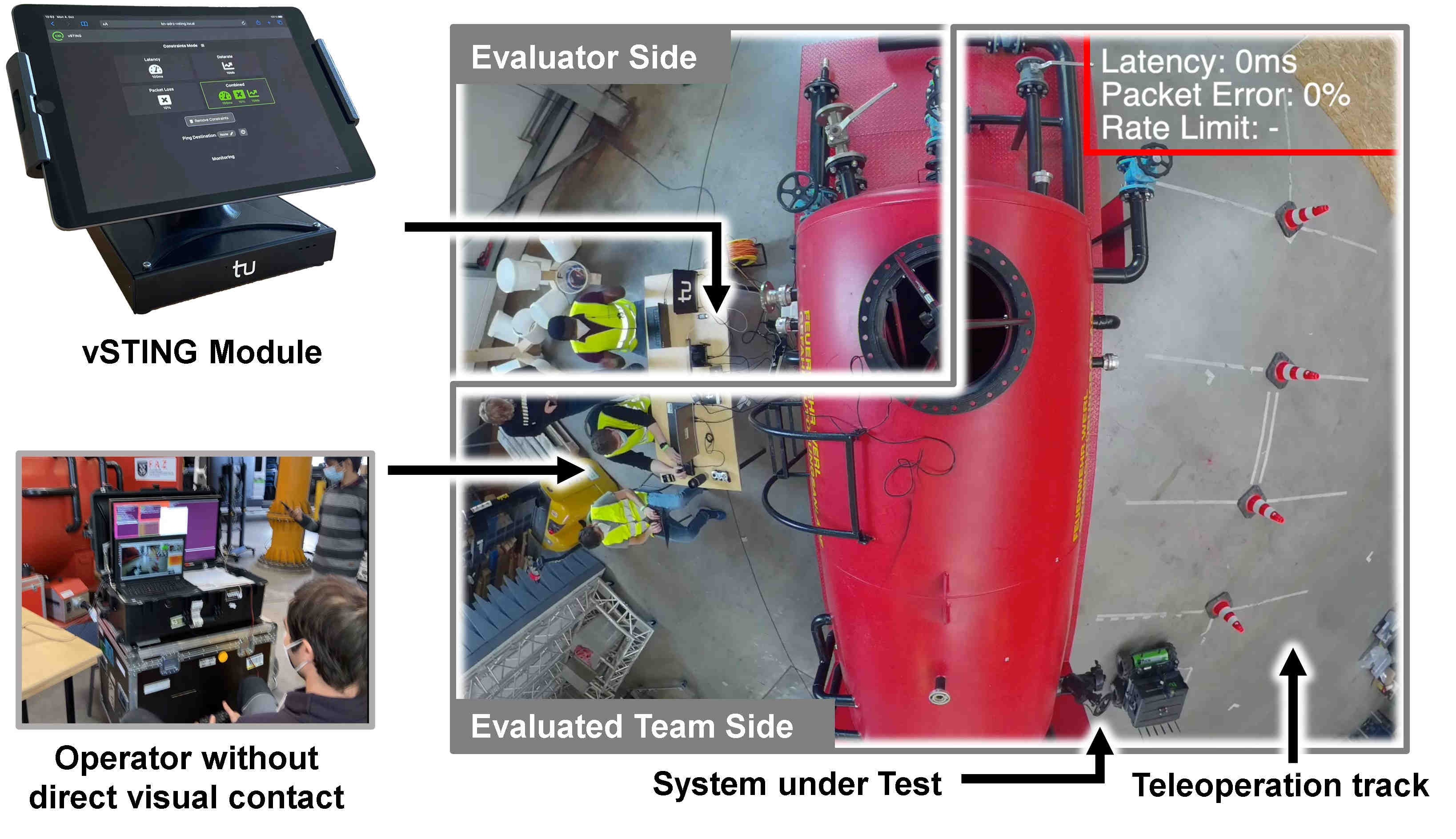}\vspace*{-2ex}
\caption{Successful application of the vSTING approach~\cite{Patchou2022}, as additional challenge at the Rescue Robotics League (German Open 2021\,\&\,2022). Teams must teleoperate the rescue robotic platforms through a course under emulated network constraints.}
\label{fig:netcomms-evaluation}
\end{figure}

Given the critical character of data acquisition, even for autonomous mobile robotic platforms, meaningful developments are also due in the network communications field. The deployment use cases in Sec.\ref{sec:deployment-use-cases} validate the relevance of multi-link communications, as it was observed that relying solely on one technology may yield a limited range and performance in rescue scenarios. Expanding on that, the robustness expected from networking solutions in real-world environments must be evaluated prior to missions through use-case-related inspection and test procedures. In the follow-up project E-DRZ, the procedures developed during the A-DRZ project, such as the vSTING \cite{Patchou2022} illustrated in Fig.~\ref{fig:netcomms-evaluation} and STING \cite{Arendt2021}, will serve as a base for repeatable assessment, validation, and certification processes of robotic systems in their ability to perform in network environments with constrained connectivity.
Furthermore, merging network communication considerations into autonomous navigation could result in communication-aware autonomous mobility with prospects of increased network robustness. A concrete and envisioned instance thereof is to restore lost network connectivity through adequate autonomous repositioning.
Finally, as the exploration of collapsed buildings likely involves multi-robot teams, the allocation of spectrum must be planned and enforced to ensure that each robot disposes of enough resources for its wireless transmission needs.

Across the board, it holds that the rescue robots must provide functionalities that are robust and can be quickly deployed. This goes down to apparent banalities such as automation of setup processes and standard system-go checks. 
It is crucial to adhere to these procedures in joint exercises, so that they are well established for RTF deployments.


\section{Conclusions and Future Work}
We presented our experiences from recent deployments of the DRZ Robotics Task Force. The DRZ RTF model based on continuous long-term close collaboration between researchers and first responders (fire fighters) has benefits for both sides. 
Testing cutting edge robotics technology, including research demonstrators, in joint exercises and real deployments enables researchers to gain better insight in end user needs and operational conditions and identify appropriate research priorities. First responders gain deeper awareness of the advanced technologies, assess the functionalities, and learn to employ them in missions. Together, they identify future potential benefits. 

The technologies involved in robot-assisted disaster response are developing very fast, and first responders normally do not have sufficient expertise. This requires the RTF model of researchers bringing in cutting-edge technology and---as the technology matures and becomes established---providing corresponding training and transfer of experience. These goals are therefore part of the DRZ long-term vision. Scenarios in the DRZ Living Lab and exercises carried out jointly by the RTF and additional first responders as well as researchers are set up to reflect the experience gathered so far and explore further challenges. 



The future research topics identified in this paper will be addressed in the project \emph{Establishing the German Rescue Robotics Center} (E-DRZ). 

\vspace*{1.5ex} 
\noindent\textit{Acknowledgment:}
This work is funded by the Federal Ministry of Education and Research (BMBF) under 
grant, 13N14860 (A-DRZ), cf. \href{https://rettungsrobotik.de}{https://rettungsrobotik.de}. 
We thank our project partners and collaborators.
\vspace*{-0.5ex} 

\bibliographystyle{IEEEtran} 
\bibliography{IEEEabrv,literatur}

\begin{thebibliography}{10}
\providecommand{\url}[1]{#1}
\csname url@rmstyle\endcsname
\providecommand{\newblock}{\relax}
\providecommand{\bibinfo}[2]{#2}
\providecommand\BIBentrySTDinterwordspacing{\spaceskip=0pt\relax}
\providecommand\BIBentryALTinterwordstretchfactor{4}
\providecommand\BIBentryALTinterwordspacing{\spaceskip=\fontdimen2\font plus
\BIBentryALTinterwordstretchfactor\fontdimen3\font minus
  \fontdimen4\font\relax}
\providecommand\BIBforeignlanguage[2]{{%
\expandafter\ifx\csname l@#1\endcsname\relax
\typeout{** WARNING: IEEEtran.bst: No hyphenation pattern has been}%
\typeout{** loaded for the language `#1'. Using the pattern for}%
\typeout{** the default language instead.}%
\else
\language=\csname l@#1\endcsname
\fi
#2}}

\bibitem{6523866}
G.-J.~M. Kruijff, F.~Pirri, M.~Gianni, P.~Papadakis, M.~Pizzoli, A.~Sinha,
  V.~Tretyakov, T.~Linder, E.~Pianese, S.~Corrao, F.~Priori, S.~Febrini, and
  S.~Angeletti, ``Rescue robots at earthquake-hit {M}irandola, {I}taly: {A}
  field report,'' in \emph{IEEE Int. Symp. on Safety, Security, and Rescue
  Robotics (SSRR)}, Nov 2012.

\bibitem{Kruijff-amatrice}
I.~{Kruijff-Korbayová}, L.~{Freda}, M.~{Gianni}, V.~{Ntouskos}, V.~{Hlaváč},
  V.~{Kubelka}, E.~{Zimmermann}, H.~{Surmann}, K.~{Dulic}, W.~{Rottner}, and
  E.~{Gissi}, ``Deployment of ground and aerial robots in earthquake-struck
  {A}matrice in {I}taly (brief report),'' in \emph{IEEE International Symposium
  on Safety, Security, and Rescue Robotics (SSRR)}, Oct 2016, pp. 278--279.

\bibitem{9597869}
I.~Kruijff{-}Korbayov{\'{a}}, R.~Grafe, N.~Heidemann, A.~Berrang, C.~Hussung,
  C.~Willms, P.~Fettke, M.~Beul, J.~Quenzel, D.~Schleich, S.~Behnke,
  J.~Tiemann, J.~G{\"{u}}ldenring, M.~Patchou, C.~Arendt, C.~Wietfeld, K.~Daun,
  M.~Schnaubelt, O.~von Stryk, A.~Lel, A.~Miller, C.~R{\"{o}}hrig,
  T.~Stra{\ss}mann, T.~Barz, S.~Soltau, F.~Kremer, S.~Rilling, R.~Haseloff,
  S.~Grobelny, A.~Leinweber, G.~Senkowski, M.~Thurow, D.~Slomma, and
  H.~Surmann, ``{German Rescue Robotics Center (DRZ): A} holistic approach for
  robotic systems assisting in emergency response,'' in \emph{IEEE
  International Symposium on Safety, Security, and Rescue Robotics (SSRR)},
  2021, pp. 138--145.

\bibitem{OperationalUseofUAS}
H.~M. Ray, R.~Singer, and N.~Ahmed, ``A review of the operational use of uas in
  public safety emergency incidents,'' in \emph{Int. Conference on Unmanned
  Aircraft Systems (ICUAS)}, 2022, pp. 922--931.

\bibitem{murphy2014disaster}
R.~Murphy, \emph{Disaster Robotics}, ser. Intelligent Robotics and Autonomous
  Agents.\hskip 1em plus 0.5em minus 0.4em\relax MIT Press, 2014.

\bibitem{tadokoro2019disaster}
S.~Tadokoro, \emph{Disaster Robotics: Results from the ImPACT Tough Robotics
  Challenge}, ser. Springer Tracts in Advanced Robotics.\hskip 1em plus 0.5em
  minus 0.4em\relax Springer, 2019.

\bibitem{DeCubber/etal:2013}
G.~De~Cubber, D.~Doroftei, D.~Serrano, K.~Chintamani, R.~Sabino, and
  S.~Ourevitch, ``The {EU-ICARUS} project: {D}eveloping assistive robotic tools
  for search and rescue operations,'' in \emph{IEEE International Symposium on
  Safety, Security, and Rescue Robotics (SSRR)}, 2013.

\bibitem{Marconi/etal:2013}
L.~Marconi, S.~Leutenegger, S.~Lynen, M.~Burri, R.~Naldi, and C.~Melchiorri,
  ``Ground and aerial robots as an aid to alpine search and rescue: Initial
  {SHERPA} outcomes,'' in \emph{IEEE International Symposium on Safety,
  Security, and Rescue Robotics (SSRR)}, 2013.

\bibitem{nardi_2019_notredame}
T.~Nardi, ``The drones and robots that helped save {N}otre {D}ame,''
  \url{https://hackaday.com/2019/04/17/the-drones-and-robots-that-helped-save-notre-dame},
  acc.2019-03-05.

\bibitem{DBLP:journals/corr/abs-1709-00587}
A.~Gawel, R.~Dub{\'{e}}, H.~Surmann, J.~I. Nieto, R.~Siegwart, and C.~Cadena,
  ``{3D} registration of aerial and ground robots for disaster response: {A}n
  evaluation of features, descriptors, and transformation estimation,''
  \emph{CoRR}, vol. abs/1709.00587, 2017.

\bibitem{Schleich:ICRA2021}
D.~Schleich and S.~Behnke, ``Search-based planning of dynamic {MAV}
  trajectories using local multiresolution state lattices,'' in \emph{{IEEE}
  Int. Conf. on Robotics and Automation (ICRA)}, 2021, pp. 7865--7871.

\bibitem{Schleich:IROS2022}
------, ``Predictive angular potential field-based obstacle avoidance for
  dynamic {UAV} flights,'' in \emph{IEEE/RSJ International Conference on
  Intelligent Robots and Systems (IROS)}, 2022.

\bibitem{Bultmann:ECMR2021}
S.~Bultmann, J.~Quenzel, and S.~Behnke, ``Real-time multi-modal semantic fusion
  on unmanned aerial vehicles,'' in \emph{10th European Conference on Mobile
  Robots (ECMR)}.\hskip 1em plus 0.5em minus 0.4em\relax {IEEE}, 2021.

\bibitem{Bultmann:RAS2022}
------, ``Real-time multi-modal semantic fusion on unmanned aerial vehicles
  with label propagation for cross-domain adaptation,'' \emph{Robotics and
  Autonomous Systems}, 2022.

\bibitem{daun2021}
K.~Daun, M.~Schnaubelt, S.~Kohlbrecher, and O.~von Stryk, ``{HectorGrapher}:
  {C}ontinuous-time lidar {SLAM} with multi-resolution signed distance function
  registration for challenging terrain,'' in \emph{IEEE Int. Symp. on Safety,
  Security, and Rescue Robotics (SSRR)}, 2021.

\bibitem{Gueldenring2020}
J.~Güldenring, P.~Gorczak, M.~Patchou, C.~Arendt, J.~Tiemann, and C.~Wietfeld,
  ``Skates: {I}nteroperable multi-connectivity communication module for
  reliable search and rescue robot operation,'' in \emph{16th International
  Conference on Wireless and Mobile Computing, Networking and Communications
  (WiMob)}, 2020, pp. 7--13.

\bibitem{9597677}
H.~Surmann, D.~Slomma, S.~Grobelny, and R.~Grafe, ``Deployment of aerial robots
  after a major fire of an industrial hall with hazardous substances, a
  report,'' in \emph{IEEE International Symposium on Safety, Security, and
  Rescue Robotics (SSRR)}, 2021, pp. 40--47.

\bibitem{9738529}
H.~Surmann, D.~Slomma, R.~Grafe, and S.~Grobelny, ``Deployment of aerial robots
  during the flood disaster in {E}rftstadt / {B}lessem in {J}uly 2021,'' in
  \emph{8th International Conference on Automation, Robotics and Applications
  (ICARA)}, 2022, pp. 97--102.

\bibitem{fabian2021}
S.~Fabian and O.~von Stryk, ``Open-source tools for efficient {ROS} and
  {ROS2}-based {2D} human-robot interface development,'' in \emph{European
  Conference on Mobile Robots (ECMR)}.\hskip 1em plus 0.5em minus 0.4em\relax
  IEEE, 2021.

\bibitem{oehler2021}
M.~Oehler and O.~von Stryk, ``A flexible framework for virtual omnidirectional
  vision to improve operator situation awareness,'' in \emph{European
  Conference on Mobile Robots (ECMR)}.\hskip 1em plus 0.5em minus 0.4em\relax
  IEEE, 2021.

\bibitem{MurphyTadokoro2019}
R.~R. Murphy and S.~Tadokoro, \emph{User Interfaces for Human-Robot Interaction
  in Field Robotics}.\hskip 1em plus 0.5em minus 0.4em\relax Springer, 2019,
  pp. 507--528.

\bibitem{Patchou2022}
M.~Patchou, J.~Tiemann, C.~Arendt, S.~Böcker, and C.~Wietfeld, ``Realtime
  wireless network emulation for evaluation of teleoperated mobile robots,'' in
  \emph{IEEE International Symposium on Safety, Security, and Rescue Robotics
  (SSRR)}, 2022.

\bibitem{Arendt2021}
C.~Arendt, M.~Patchou, S.~Böcker, J.~Tiemann, and C.~Wietfeld, ``Pushing the
  limits: Resilience testing for mission-critical machine-type communication,''
  in \emph{IEEE 94th Vehicular Technology Conference (VTC2021-Fall)}, 2021.

\end{thebibliography}

\end{document}